\documentclass[runningheads]{llncs}
\usepackage{graphicx}
\usepackage{comment}
\usepackage{amsmath,amssymb} 
\usepackage{color}
\usepackage{capt-of}

\newcommand*\diff{\mathop{}\!\mathrm{d}}

\begin{document}
\pagestyle{headings}
\mainmatter

\title{Learning to hash with semantic similarity metrics and empirical KL divergence} 

\author{Heikki Arponen and Tom E. Bishop}
\institute{Intuition Machines, Inc.}

\maketitle

\begin{abstract}
 Learning to hash is an efficient paradigm for exact and approximate nearest neighbor search from massive databases. Binary hash codes are typically extracted from an image by rounding output features from a CNN, which is trained on a supervised binary similar/ dissimilar task. Drawbacks of this approach are: (i) resulting codes do not necessarily capture semantic similarity of the input data (ii) rounding results in information loss, manifesting as decreased retrieval performance and (iii) Using only class-wise similarity as a target can lead to trivial solutions, simply encoding classifier outputs rather than learning more intricate relations, which is not detected by most performance metrics. We overcome (i) via a novel loss function encouraging the relative hash code distances of learned features to match those derived from their targets. We address (ii) via a differentiable estimate of the KL divergence between network outputs and a binary target distribution, resulting in minimal information loss when the features are rounded to binary. Finally, we resolve (iii) by focusing on a hierarchical precision metric. Efficiency of the methods is demonstrated with semantic image retrieval on the CIFAR-100, ImageNet and Conceptual Captions datasets, using similarities inferred from the WordNet label hierarchy or sentence embeddings.
\end{abstract}

\section{Introduction}\label{intro}


As image databases grow ever larger, the importance of performing efficient nearest neighbor (NN) based similarity search has increased. One effective paradigm for this is \emph{learning to hash}, whose objective is to learn a mapping from images to binary hash codes, such that a query image's code is close to semantically similar images in a pre-computed database (when measured e.g. in terms of Hamming distance). In addition to offering fast exact NN search by using e.g. Multi-Index Hashing \cite{norouzi2013fast}, the binary hash codes also result in considerably lower storage requirements. A good review on learning to hash is given by Wang et al.~\cite{Wang:2016up}.

Two classes of hashing for retrieval are \textit{Data-independent} methods such as the classic (LSH) \cite{Gionis:1999wf}, and \textit{Data-dependent} (either supervised or unsupervised) methods, which learn hash codes tailored to the datasets in question and have shown considerably better performance. 
\emph{Deep} hashing methods use Deep Neural Networks (DNNs) as the hash function, by binarizing network outputs. They have shown major advantages over traditional hashing in the supervised \cite{Cao:2017vm}, \cite{Liong:2015ig}, \cite{Jiang:2017tf}, \cite{Li:2018tj}, \cite{Li:2018tj} and unsupervised \cite{Jin:2018vo} settings.

Most supervised hashing techniques use only pairwise binary similarity, i.e. only images of the same class are considered in any way similar. However, such a measure is too crude to properly learn deeper semantic relationships in data. As excellently illustrated in \cite{Barz:2018vt}, models trained with pairwise similarities need not learn anything of inter-class relations: a cat-dog pair has the same distances as cat-airplane. We demonstrate and measure this effect in our results.

Another key issue of most deep hashing approaches is the separation into (i) learning a continuous embedding via backpropagation (ii) subsequent thresholding of these into hash codes. Hence the differentiable optimization and hashing objectives are not aligned; the model is not forced to learn exactly binary outputs, and information loss occurs when performing the thresholding operation.

A third and probably the most critical problem with typical evaluation of the supervised learning to hash paradigm was exposed in Sablayrolles et al. \cite{Sablayrolles:2016uu}: a trivial classifier solution that simply outputs a one-hot class prediction as a hash code usually results in state-of-the-art performance when measured with traditional retrieval metrics, such as the mean average precision (mAP), while performing poorly on novel datasets with unseen labels. Note that this doesn't mean that methods reporting only mAP would necessarily be bad; just that e.g. mAP alone is a grossly inadequate measure for retrieval performance. 

\paragraph{Our main contributions are as follows:} 
\begin{enumerate}
    \item We address the limits of \emph{pairwise binary similarity} by using more nuanced similarity metrics: using semantic distances between labels or other supervisory signals, such as captions. Specifically, rather than directly using ``image x has label y" we consider relations such as ``image $x_1$ is as similar to $x_2$ as label $y_1$ is similar to $y_2$". For a batch of B images, the target is then a BxB size matrix with elements corresponding to a semantic similarity distance between the corresponding labels. Hence we are using information between all examples in the batch, instead of e.g. a triplet based approach, which may require expensive mining schemes. This results in superior performance compared to binary similarity, with an insignificant computational overhead.
    \item We resolve the \emph{hash code binarization} problem by introducing a novel loss function based on the differentiable Kozachenko-Leonenko estimator of the KL divergence \cite{wang2006nearest}. We minimize the KL divergence between a balanced binary (Beta) target distribution and the continuous valued network outputs. 
    Note that this is an estimate for the true KL divergence between a given target distribution and distribution of $Y = f(X)$, $X \sim P_{data}$ for a network $f$. This regularizes network outputs towards binary values, with in minimal information loss after thresholding. Additionally, these hash codes will maximally utilize the hash code space since they are uniformly distributed, which is crucial in avoiding the trivial solution problem, as well as in guaranteeing efficient retrieval when using e.g. the Multi-Index Hashing approach \cite{norouzi2013fast}.
    \item We measure the performance of our methods by focusing on the mean average hierarchical precision (mAHP) \cite{Deng:2011fw} in addition to mAP and accuracy metrics, that have been shown to be highly unreliable for supervised retrieval. The mAHP measure will take into account the similarity between the query and retrieval results, and will therefore be a more reliable retrieval metric. Specifically, our experiments demonstrate that models trained only with a classification loss may indeed suffer from low mAHP while seemingly performing very well in terms of mAP. While mAHP is undoubtedly a better retrieval metric than mAP, the ultimate test of any retrieval model is how it performs with completely \emph{unseen} data. We therefore also measure our method's performance in the Zero Shot Hashing (ZSH) setting \cite{shen2019scalable}, where a model is trained with the 1000 label ILSVRC2012 dataset, and tested with the full ImageNet dataset consisting solely of classes \emph{unseen} during training.
\end{enumerate}

We test our method and obtain state-of-the-art retrieval results in terms of mAHP on the CIFAR-100\cite{krizhevsky2009learning} and ImageNet (ILSVCR2012)\cite{ILSVRC15} datasets. 
We experiment with (i) using the WordNet\cite{wordnet} hierarchy to define a non-binary semantic similarity metric for the labels; and (ii) sentence embeddings computed on the class labels' WordNet descriptions by using a version of BERT\cite{devlin2018bert} fine-tuned on the MRPC corpus\cite{cer2017semeval}. 

We extend this method to work with weakly-labeled captions in the form of Google's Conceptual Captions (CC)\cite{sharma-etal-2018-conceptual} dataset, where we generate sentence embeddings with BERT, and use these embeddings to define a distance matrix, enabling retrieval of images with nuanced semantic content beyond simple class labels. 
In addition, we test our methods in the ZSH setting with the full 21k-class ImageNet unseen label dataset, and demonstrate that our method is effective in a real world content retrieval setting.

\section{Related work}
In Unsupervised Semantic Deep Hashing (USDH, \cite{Jin:2018vo}), semantic relations are obtained by looking at embeddings on a pre-trained VGG model on ImageNet. The goal of the semantic loss here is simply to minimize the distance between binarized hash codes and their pre-trained embeddings, i.e. neighbors in hashing space are neighbors in pre-trained feature space. 
This is somewhat similar to our notion of semantic similarity except that they use a pre-trained embedding instead of a labeled semantic hierarchy of relations.
Some works \cite{Zhe:2019vi,Zhe:2018wg} consider class-wise Deep hashing, in which a clustering-like operation is used to form a loss between samples from the same class and, in \cite{Zhe:2019vi}, across levels from the hierarchy.  In our method, we do not require explicit targets to be learned by the network, only relative semantic distances to be supplied as targets.

Recently \cite{Barz:2018vt} explored image retrieval (without hashing) using semantic hierarchies to design an embedding space, in a two-step process.
Firstly, they directly find embedding vectors of the class labels on a unit hypersphere, via a linear algebra-based approach, such that distances of these embeddings are similar to the supplied hierarchical similarity.
In Stage 2, they train a standard CNN encoder model to regress images towards these embedding vectors. 
We make use of hierarchical relational distances in a similar way to constrain our embeddings. We will however not regress towards any pre-learned fixed target embeddings, but instead require that the distances between the neural network outputs will match the distances inferred from the target similarities. Also, unlike our work, \cite{Barz:2018vt} use only continuous representations and require an embedding dimension equal to the number of classes, whereas we learn compact binary hash codes of any dimension, enabling fast queries from huge databases.

In \cite{sablayrolles2018spreading}, embedding vectors on a unit hypersphere are also used. They use the Kozachenko-Leononenko nearest neighbor based estimator to maximize entropy on this hypersphere to encourage uniformity of the network outputs. We also use this estimation method, but to minimize the Kullback-Leibler divergence between the (empirical) distribution of network outputs and a target distribution, which in our case is a near-binary one. Also, they consider only a binary similarity matrix based on nearest neighbors in the input space, using a triplet loss. They also do not consider hashing.

We also consider how semantic relations in language models apply to guiding similarity in the image domain. Pre-trained word embeddings have found wide adoption in many NLP tasks \cite{Goldberg:2015vc}; however, their suitability for various transfer learning tasks depends heavily on the particular domain and training dataset. Therefore, considering combinations of different word embeddings (``meta-embeddings'') \cite{Yin:2015wy},  \cite{Coates:2018uq} or sentence embeddings is a promising direction.
%
%
In DeViSE \cite{Frome:2013ux}, a language model and visual model are combined, in order to transfer knowledge from the textual domain to the visual domain, starting from pre-trained models in each domain, and attempting to minimize distances in embedding space between text and image representations. 
Rather than trying to directly learn a mapping from images to word embeddings, we can use our relaxed similarity distance matching objective to align these two domains.

\section{Learning binary hash codes while preserving semantic similarity}
Suppose we have a dataset of images $\left\{x_n\right\}_{n=1}^N$ and corresponding targets $y_n$, which could be labels, attributes, captions etc. We wish to learn \emph{useful representations} $\hat z_n = f\left(x_n ; \theta  \right)$ by backpropagation by using a deep neural network $f$ with weights $\theta$, but so that the network outputs $\hat z$ are \emph{binary valued}, for use as hash codes in fast hash-based queries from massive databases. Clearly just thresholding the network output to generate binary vectors would not work, because the sign function is not differentiable. One proposed solution in \cite{he2018hashing} is to use a ``soft sign" function  with adjustable scaling. This would however not guarantee uniformity, nor that the actual values would be near binary.

Usefulness of the representation is in our case defined by semantic similarity: network outputs of semantically similar images should be close to each other when measured in Hamming distance. In the optimal case, there would be a continuous measure of semantic similarity based on the targets $y$, which the network would learn to reproduce as accurately as possible in the Hamming distances between the network outputs. These are the two old problems in learning to hash, for which we propose two novel solutions, in sections \ref{LearningSemSimFromData} and \ref{MinEmpKL}.

\subsection{Overall Methodology}
\begin{figure}[h]
\begin{center}
\includegraphics[keepaspectratio=true,scale=0.29]{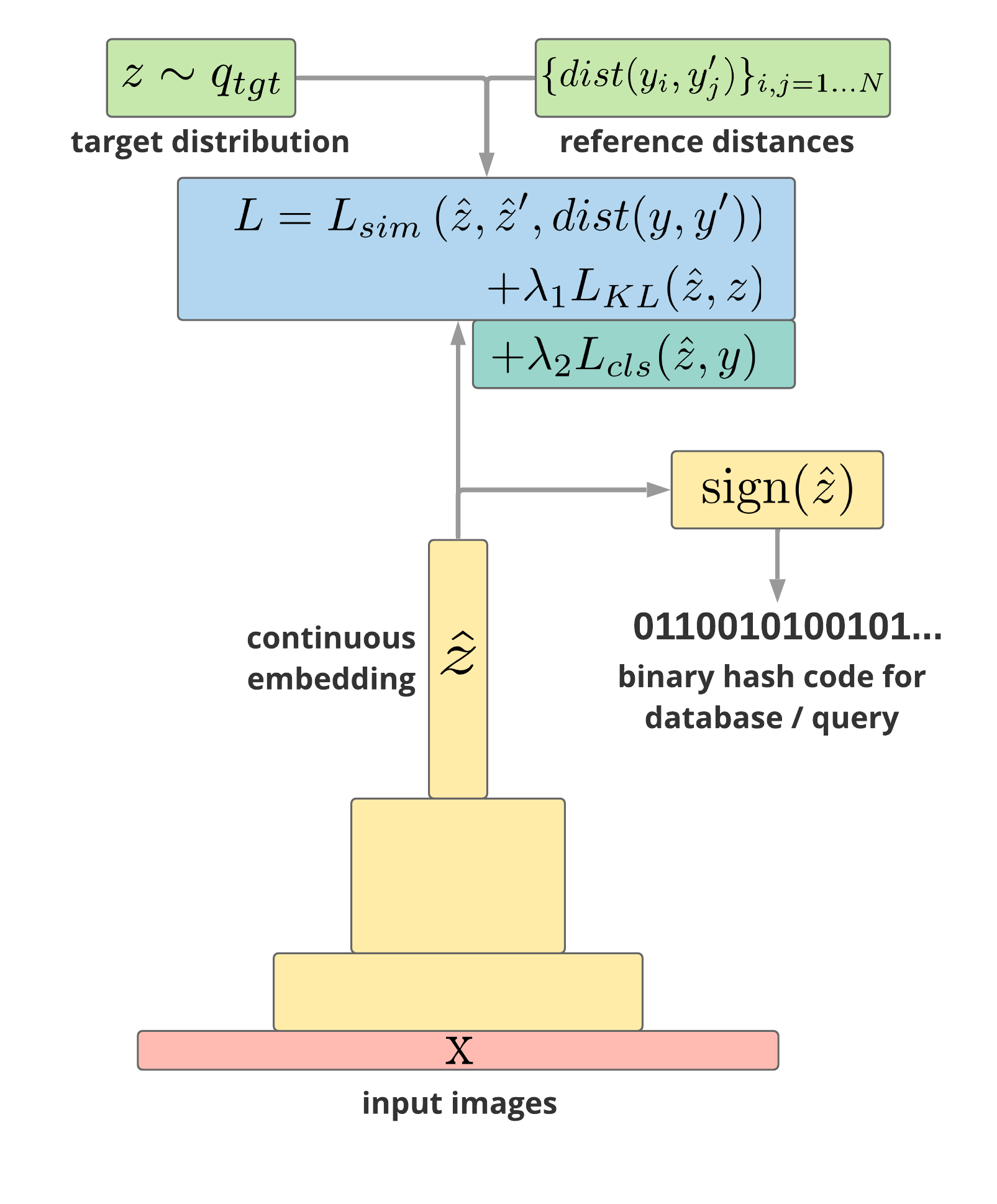}
\includegraphics[keepaspectratio=true,scale=0.25]{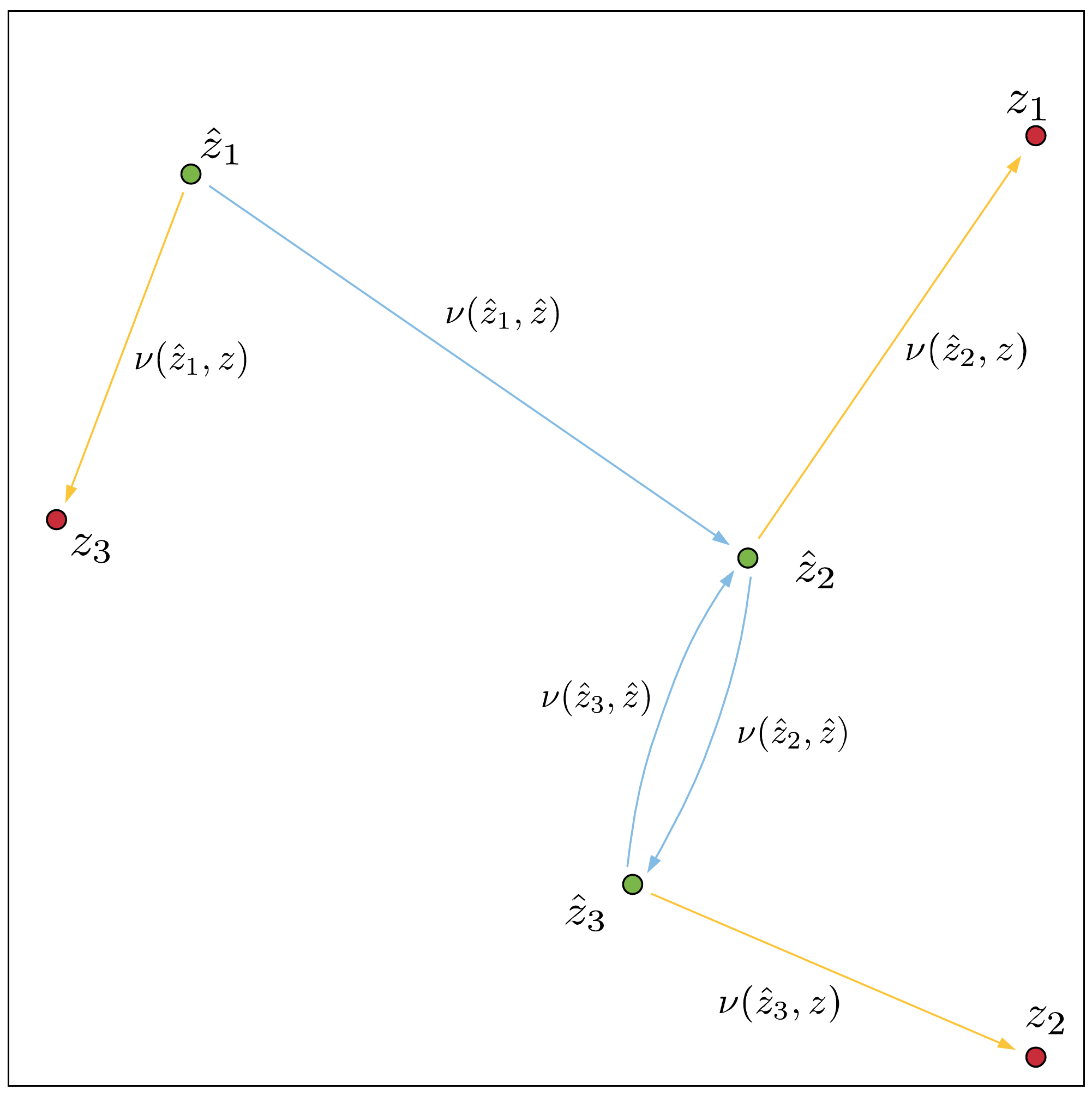}
\end{center}
\caption{\emph{Left} Network overview. A minibatch of images $x_b$, $b=1, \ldots, B$ are fed into an encoder. Output $\hat z_b$ is a continuous valued embedding of e.g. dimension 64. We then match the Manhattan distances between outputs $|\hat z_b - \hat z_{b'}|$ with the distances $dist(y_b, y_{b'})$ derived from the labels. Actual binary codes are not used during training, but instead an empirical KL loss is added to regularize the $\hat z$ towards binary values specified by a binary target distribution $q_{tgt}$.  The combined loss function is optimized until convergence, after which we extract the hash codes for retrieval by thresholding $\hat z$, which are now already near binary values.
\emph{Right} Visualization of the k-Nearest Neighbor estimation of KL divergence for three sample points both from the latent space (green) and the target distribution (red). Meaning of symbols is the same as in main text. Arrowheads point \emph{to} the nearest neighbor between codes $\hat z$ and samples from target distribution $z$ (yellow) and between the codes themselves (blue).}
\label{Network}
\end{figure}

We train a CNN with a bottleneck output layer that we can binarize to produce both database and query codes for image retrieval; see Figure~\ref{Network}~(a). A target distribution is used to constrain the continuous embedding to be approximately balanced and near-binary.
Our total loss function is defined as
\begin{align}
    L &= L_{sim} + \lambda_1 L_{KL} + \lambda_2 L_{cls}, \label{eqLoss}
\end{align}
where $L_{cls}$ is a standard classification cross-entropy loss in most experiments (replaced by a regression loss for the Conceptual Captions dataset). 
$L_{sim}$ and $L_{KL}$ refer to the similarity and KL divergence loss terms defined in the next sections, and $\lambda_1$ and $\lambda_2$ are hyperparameters, set to scale the various loss contributions approximately equally, which occurs at $\lambda_1 = \lambda_2=0.01$.

\subsection{Learning semantic similarity from data}\label{LearningSemSimFromData}
We assume that it is possible to infer the semantic similarity between two images $x$, $x'$ from the corresponding targets $y$, $y'$ by a distance measure $d(y, y')$. The most common approach is to define the measure by using labels $y$ and defining $d(y, y') = 0$ when the labels are the same, and $d(y, y') = 1$ when they are different. In a worst case scenario, using such a binary optimization criterion could lead the model to learn a unique hash code corresponding to each label, which would make the model effectively a classifier. Even worse, the model would still seem very strong when measured in terms of the mAP (mean-average precision) score, as was excellently explained in e.g. \cite{Ding:2018wj}. We will therefore choose to work with datasets from which we can infer multiple or continuous levels of similarity.

While it would be possible to first learn embeddings for the targets $y$ as was done in e.g. \cite{Barz:2018vt} and then train the network to regress to these embeddings, we will instead only use the inferred distances and let the network adapt to whichever outputs $\hat z$ are most natural, while respecting the inferred distances. 

Suppose that the image and target pairs are denoted as $(x_b, y_b)$ for minibatch index $b = 1, \ldots, B$ and denote the elements of the inferred similarity matrix between targets $y_b$ and $y_{b'}$ as $d_{b b'} = d\left(y_b, y_{b'}\right)$. We seek to match this distance with the Manhattan distance $\left\Vert \hat z_b - \hat z_{b'}\right\Vert_M$ between the corresponding neural network outputs $\hat z_b = f (x_b; \theta)$. We define a loss term
\begin{equation}
    L_{sim} = \sum\limits_{b, b'=1}^{B} \left\vert \frac{1}{\tau_z} \left\Vert \hat z_b - \hat z_{b'}\right\Vert_M - \frac{1}{\tau_y}d_{b b'}\right\vert w_{b b'} ,
\end{equation}
where the term $w_{b b'}$ is an additional weight, which is used to emphasize similar example pairs (e.g. cat-dog) more than distant ones (e.g. cat-airplane). We use such a weighting because we are mostly interested in retrieving images that are very similar to query images. Furthermore, we are not interested as much in learning the absolute distances between examples, but instead the relative ones. Hence we have added normalizing terms $\tau_z = \sum_{b, b'=1}^{B} \left\Vert \hat z_b - \hat z_{b'}\right\Vert_M$ and $\tau_y = \sum_{b, b'=1}^{B} d\left(y_b, y_{b'}\right)$, which render the loss term scale invariant both in terms of the network outputs and target distances. This will enable the network to learn outputs as flexibly as possibly, while still respecting the relative distances learned from the targets. We have observed both these schemes to be helpful during learning. We use a slowly decaying form for the weight, $w_{b b'} = \gamma ^\rho / \left(\gamma + d_{b b'} \right)^\rho$, with parameter values $\gamma = 0.1$ and $\rho=2$.
    
\subsection{Minimizing empirical KL divergence}\label{MinEmpKL}
We are interested in extracting binary hash codes out of the network outputs $\hat z$ to facilitate fast queries from massive databases. A simple approach to doing this after training has been completed would be to simply round the $\hat z$ to a binary value (or equivalently to take the sign). This would however lead to information loss because the semantic similarity will not be preserved, and because values near zero would be assigned essentially randomly to binary values. Additionally, there are no guarantees that the hash codes would be utilized efficiently. Therefore, we want to impose that the distribution of outputs will be (nearly) binary and maximally uniform, while retaining differentiability of the loss function to allow learning by backpropagation. This way the information loss due to rounding will be minimal, since the $\hat z$ values will already be close to binary. This would further ensure that images within a given class don't share a unique hash code but will also be spread out. We could achieve this in principle by minimizing the Kullback-Leibler divergence between the distribution of the network outputs $p$ and the target distribution $q$ (with standard abuse of notation), defined as:
\begin{align}
KL(p||q) &= \int \!\! \diff z p(z) \log\left(\frac{p(z)}{q(z)}\right) \nonumber\\
         &= -\int \!\! \diff z p(z) \log (q(z)) + \int \!\! \diff z p(z) \log (p(z))\nonumber\\
         &= H(p, q) - H(p), \label{kl_analytical}
\end{align}
where on the last line we have split the KL divergence into cross-entropy and entropy terms. The KL divergence attains the minimum value of zero if (and only if) $p = q$. Analytical solution of such an objective is of course intractable in general, so we resort to instead minimizing a k-nearest neighbor Kozachenko-Leonenko empirical estimate of the KL divergence (see \cite{wang2006nearest}). Our empirical loss term for achieving this is defined as
\begin{align}
    L_{KL} &= \frac{1}{B} \sum\limits_{b=1}^B \left[\log \left(\nu(\hat z_b ; z)\right) - \log\left(\nu(\hat z_b ; \hat z)\right)  \right] \nonumber\\
    &\doteq \widehat H(p, q) - \widehat H(p),
    \label{kl_estimate_eq}
\end{align}
where $\nu(\hat z_b ; z)$ denotes the distance of $\hat z_b$ to a nearest vector $z_{b'}$, and $z$ is a sample (of e.g. size $B$) of vectors from a target distribution. We have again performed a split into empirical cross-entropy and entropy terms on the second line, corresponding to the analytical expressions in Eq. (\ref{kl_analytical}). It is now easy to see intuitively how such a loss function will be minimized: the samples $\hat z_b \sim p(z)$ drawn from the ``model distribution" will need to be as close as possible to the samples $z_b' \sim q(z)$ drawn from the target distribution, while at the same time making sure that all the $\hat z_b$ won't ``mode collapse" to a same $z_b'$ by maximizing the inter-distance between the $\hat z_b$, i.e. the empirical entropy of $p(z)$. 
%
%
%

\subsection{Distribution matching via the Kozachenko-Leonenko KL divergence estimator}
Our novel KL-based loss term has been used to help with hash-code binarization, but could impose an arbitrary distribution. We give some more intuitive exposition of its function in Figure~\ref{Network}~(b), where we represent graphically how this empirical Kozachenko-Leonenko estimator forms an estimate of the Kullback-Leibler divergence between an the k-Nearest Neighbors of an observed distribution and a target distribution. Effectively, it consists of ensuring the mean \emph{within}-distribution (observed--observed) nearest neighbor distance of each sample is similar to that of the closest \emph{across}-distribution (observed--target) distance.

\section{Experiments}
We implement our method in PyTorch\cite{paszke2017automatic} with mixed precision training distributed across up to 64 NVIDIA Tesla V100 GPUs. 
We study ablations of models trained with the different losses in Eq.~\ref{eqLoss}, denoted as ``SIM", ``KL" and ``CLASS" and combinations thereof, across 3 datasets.

For CIFAR-100 (Section~\ref{sec:CIFAR}) and ImageNet (Section~\ref{sec:Imagenet}), to define $d(y,y')$, we use NLTK's \cite{bird2009natural} implementation of the Wu-Palmer Similarity metric (WUP) \cite{martin2009speech}, which defines similarity between class labels as the shortest number of edges between them in the WordNet\cite{wordnet} graph, weighted by the distance in the hierarchy. For ImageNet, we also consider using sentence embeddings computed on the class labels' WordNet descriptions by using a version of BERT\cite{devlin2018bert} finetuned on the MRPC corpus\cite{cer2017semeval}.
For all experiments we included comparisons with available prior work that included mAHP in their results.

In Section~\ref{sec:CC}, we train a model on the Conceptual Captions dataset, also using the BERT embeddings of the images' captions as a reference similarity metric.
We also report in Section \ref{sec:ZSH} some results in a Zero Shot Hashing setting when retrieving from a dataset with totally unseen classes, demonstrating the generalization capabilities of our method.

As discussed in the introduction, using mAP to measure retrieval performance is imprecise: a perfect classifier can yield a perfect mAP score, while having poorly distributed hash codes and no notion of similarity between classes. We therefore focus on the mAHP score, and show how a network trained with a classification loss only has a high mAP, but poor mAHP score.

\begin{table}[t]
\begin{center}
\begin{tabular}{lccc}
\\ \hline
\multicolumn{1}{c}{\bf Code length}  &\multicolumn{1}{c}{\bf mAHP}  &\multicolumn{1}{c}{\bf mAP} &\multicolumn{1}{c}{\bf Class acc.}
\\ \hline
16 bits     &    0.7478           & 0.3577             & 65.65\%   \\ 
32 bits     &    0.8202           & 0.5114             & 65.00\%   \\ 
64 bits     &    0.8690         & 0.6514        & 70.79\% \\
128 bits    &    \textbf{0.8760}  & \textbf{0.6857}    & 70.29\% \\ 
\hline
\end{tabular}
\caption{Effect of hash code length on CIFAR-100 for $L_{sim} + L_{kl}$\label{cifar2}}
\end{center}
\end{table}

\subsection{CIFAR-100}\label{sec:CIFAR}

\begin{table*}[t]
\begin{center}
\begin{tabular}{l|l||l||l|l}
\hline
\textbf{Method}                                          & \textbf{mAHP}     & \textbf{mAHP bin}    & \textbf{mAP}      & \textbf{Class acc.}   \\ 

\hline
\textit{Others}                                          & {}         & {}                & {}               & {}       \\
DeViSE \cite{Frome:2013ux}          & 0.7348     & -   & 0.5016          & 74.66\%    \\
Centre Loss, \cite{Wen:2016jx}      & 0.6815     & -   & 0.4153          & 75.18\%    \\
Label Embedding, \cite{Sun:2017wl}  & 0.7950     & -   & 0.6202          & \textbf{76.96\%}    \\
\cite{Barz:2018vt}, $L_{CORR}$      & 0.8290     & -   & 0.5900          & 75.03\%    \\
\cite{Barz:2018vt}, $L_{CORR+CLS}$  & 0.8329     & -   & 0.6107          & 76.60\%    \\
\cite{Zhe:2019vi}\textsuperscript\ddag  & (0.8667)     & -   & (0.8259)          & \text{n/a} \\

\hline
\textit{Ours}               & {}         & {}                & {}               & {}       \\
SIM          & 0.8658    & 0.8431    & 0.5983     & 68.92\%  \\
CLASS        & 0.8653    & 0.8380    & 0.6756     & 71.20\%  \\
KL-CLASS     & 0.5850    & 0.5413    & 0.2915     & 62.29\%  \\
SIM-KL       & 0.8655    & 0.8690    & 0.6514     & 68.54\%  \\
SIM-KL-CLASS & 0.8916    & 0.8608    & 0.6512     & 71.99\%  \\

\hline
\end{tabular}
\caption{CIFAR-100. ``SIM" refers to using the similarity loss, ``KL" to the empirical KL divergence loss, and ``CLASS" to the cross entropy loss. When ``CLASS" is absent, we use the hash codes simply as features for an additional classifier and do not backpropagate through the rest of the network. ``mAHP bin" refers to mAHP as measured from the binarized hash codes. Note that our mAP results are also computed from the binary valued codes. All our results in the table are with 64-dimensional hash codes. Other results are from \cite{Barz:2018vt}. \ddag All mAP and mAHP computed @250, except \cite{Zhe:2019vi}  uses mAP and mAHP @2500, so not really comparable; included for completeness.  \label{cifar_table}}
\end{center}
\end{table*}

We used the Resnet-110w model architecture as in \cite{Barz:2018vt}, where the top fully connected layer is replaced to return embeddings at the size of the desired hash length. We also added a small classification head, which is detached from the rest of the network when not using  $L_{cls}$ as an additional task. We used the Adam optimizer with a learning rate $0.0005$ and trained the model for 500 epochs on a single V100 GPU. We used $10^{-4}$ weight decay and a batch size of 512. 

We see in Table \ref{cifar_table} how our method improves considerably over previous comparable methods for both mAHP and mAP scores, while using only 64-dimensional binary valued codes. Note especially how binary mAHP does not drop compared to float mAHP for SIM-KL (in fact it increases within error margins), whereas there's a noticeable drop without the KL loss. There is a drop in SIM-KL-CLASS, which could be due to the additional classification loss hampering the effects of the KL loss. Note that when $L_{cls}$ is absent, the codes/ network outputs are used as fixed representations for classification, i.e. gradients are not propagated through the main hashing network.

We have also provided a study on the effect of the code length (network output dimension) on mAHP and mAP scores, validating that longer codes 
are in general better for retrieval, but also that the difference from 64 to 128 dimensions is not substantial, at least for CIFAR-100.

\subsection{ImageNet}\label{sec:Imagenet}

\begin{table*}[t]
\centering
\begin{tabular}{l|c||c||c|c|c}
\hline
\textbf{Method} & \textbf{mAHP}     & \textbf{mAHP}    & \textbf{mAP}      & \textbf{Classification}  & \textbf{Binary} \\ 
& \textbf{@250}     & \textbf{@250 bin}     & \textbf{@250}       &  \textbf{Accuracy} &  \textbf{Entropy}\\ 
\hline
\textit{Others:}                             & {}         & {}                & {}               & {}      &     \\
Centre Loss, \cite{Wen:2016jx}             & 0.4094     & -                 & 0.1285           & 70.05\%   & -  \\
Label Embedding, \cite{Sun:2017wl}         & 0.4769     & -                 & 0.2683           & 70.94\%   & -  \\
\cite{Barz:2018vt} $L_{CORR}$ (1000 dim float)  & 0.7902     & -            & 0.3037           & 48.97\%   & -  \\
\cite{Barz:2018vt} $L_{CORR+CLS}$  (1000-d float)& 0.8242     & -         & 0.4508           & 69.18\%    & - \\
\hline
\textit{Ours}: Trivial One-hot solution             & 0.4389     & -                 & 0.7547           & 75.40\%   & 16.31  \\
\hline
\textit{Ours: WUP, 64-d codes:}       & {}         & {}            & {}               & {}       \\
SIM-KL      & 0.8620     & 0.8306            & 0.4500           & 59.41 \%    & 19.37  \\
SIM-KL-CLASS& 0.8952     & \textbf{0.8818}   & 0.6508           & 70.44 \%   &  19.41  \\
KL-CLASS            & 0.8903     & 0.7956   & 0.7075           & 74.19 \%   &  20.42  \\
CLASS               & 0.8872     & 0.7424   & 0.6347           & 74.35 \%   &  16.49  \\

\textit{Ours, BERT, 64 dim codes:}       & {}         & {}            & {}               & {}       \\
SIM-KL            & 0.9286   & 0.9274           & 0.6982    & 68.75 \%   & 19.78    \\ 
SIM-KL-CLASS      & 0.9393   & \textbf{0.9399}  & \textbf{0.7611} & 72.84 \%  & 19.77    \\
KL-CLASS            & 0.9316     & 0.9262   & 0.7379  & 74.94 \%  &  20.43   \\
CLASS               & 0.9186     & 0.8858   & 0.6118  & \textbf{75.46} \%  & 16.40    \\
\hline
\end{tabular}
\caption{Results for ImageNet (ILSCVR2012) dataset. We report mAHP@250 for float and binary (bin) outputs. We emphasize that the mAHP scores should not be compared \emph{between} the WUP and BERT embeddings, as the value depends on the actual distance instead of simply ranking. It is however clear from the mAP and Accuracy that the BERT embeddings are of much higher quality than the WUP ones. ``\emph{Others}" are results from \cite{Barz:2018vt}. Best model overall is the ``SIM-KL-CLASS" with BERT embeddings, when measured in terms of the binary mAHP. Note that when the KL loss is not present, the Binary Entropy is low and corresponds roughly to only around 1000 unique hash codes for the entire database. Note also how the Trivial solution consisting of a "hash code" of class predictions yields a SotA mAP score, but a poor mAHP score.\label{imagenet_table}}
\end{table*}

We also used the ImageNet Large Scale Visual Recognition Challenge (ILSVRC) 2012 dataset. We used ResNet50 \cite{he2016deep} with slight modifications as in \cite{he2019bag}, where output is now equal to the hash code dimension. We again added a small classification head detached from the rest of the network when not learning using the classification loss $L_{cls}$. We use a standard ImageNet distributed training scheme, using SGD with Nesterov momentum with a learning rate $0.05$ scaled by the number of GPUs. We conducted all training with 32 GPUs and for a total of 320 epochs, decaying the learning rate by $0.1$ every 100 epochs. We used $10^{-4}$ weight decay and a batch size of $256$ with mixed precision training. We also use sentence embeddings computed on the class labels' WordNet descriptions by using a version of BERT\cite{devlin2018bert} finetuned on the MRPC corpus\cite{cer2017semeval}. 

Results together with an ablation study are reported in Table \ref{imagenet_table}. Perhaps the most important entry in the table is the ``Trivial solution", where the output code is just the 1000 dimensional class prediction vector. It was observed already in \cite{Sablayrolles:2016uu} that such trivial solutions seem to be excellent retrieval models when measured in terms of mAP. We see however that the mAHP score is very low, as well as obviously the binary entropy, given there are only 1000 unique, equidistant ``hash codes". \emph{Therefore the retrieval mAP score is not a good retrieval metric in the supervised learning scheme.} The best model is SIM-KL-CLASS with BERT embeddings. Note especially how adding the KL loss to a classification model results in only a small difference between float and binary mAHP. Also, adding a classification loss improves mAP and classification accuracy, but does not have a substantial effect on mAHP. We note especially that the BERT sentence embeddings are qualitatively much better than the WUP similarity in terms of mAP and Accuracy. Note however that the mAHP scores between WUP and BERT are not comparable, since mAHP values depend on the distance values. 

We also report the Binary Entropy as a measure of the diversity of the binary hash codes, which is estimated by the Kozachenko-Leonenko estimator of entropy\cite{wang2006nearest} (the entropy term in Eq. \ref{kl_estimate_eq}). The value 16.4 for the classification-only trained networks is especially low, corresponding to only around 1000 unique but uniformly distributed hash codes (although in reality there are probably more than 1000 dense clusters of codes). Completely uniformly distributed sampled codes in 64 dimensions would amount to entropy of 19.93, which is lower than the highest values in the table. This is simply because the the network learns to not cluster the codes close to each other, whereas samples drawn from a truly uniform distribution can be arbitrarily close to each other.

\begin{figure}[t]
\begin{center}
\includegraphics[keepaspectratio=true,scale=0.35]{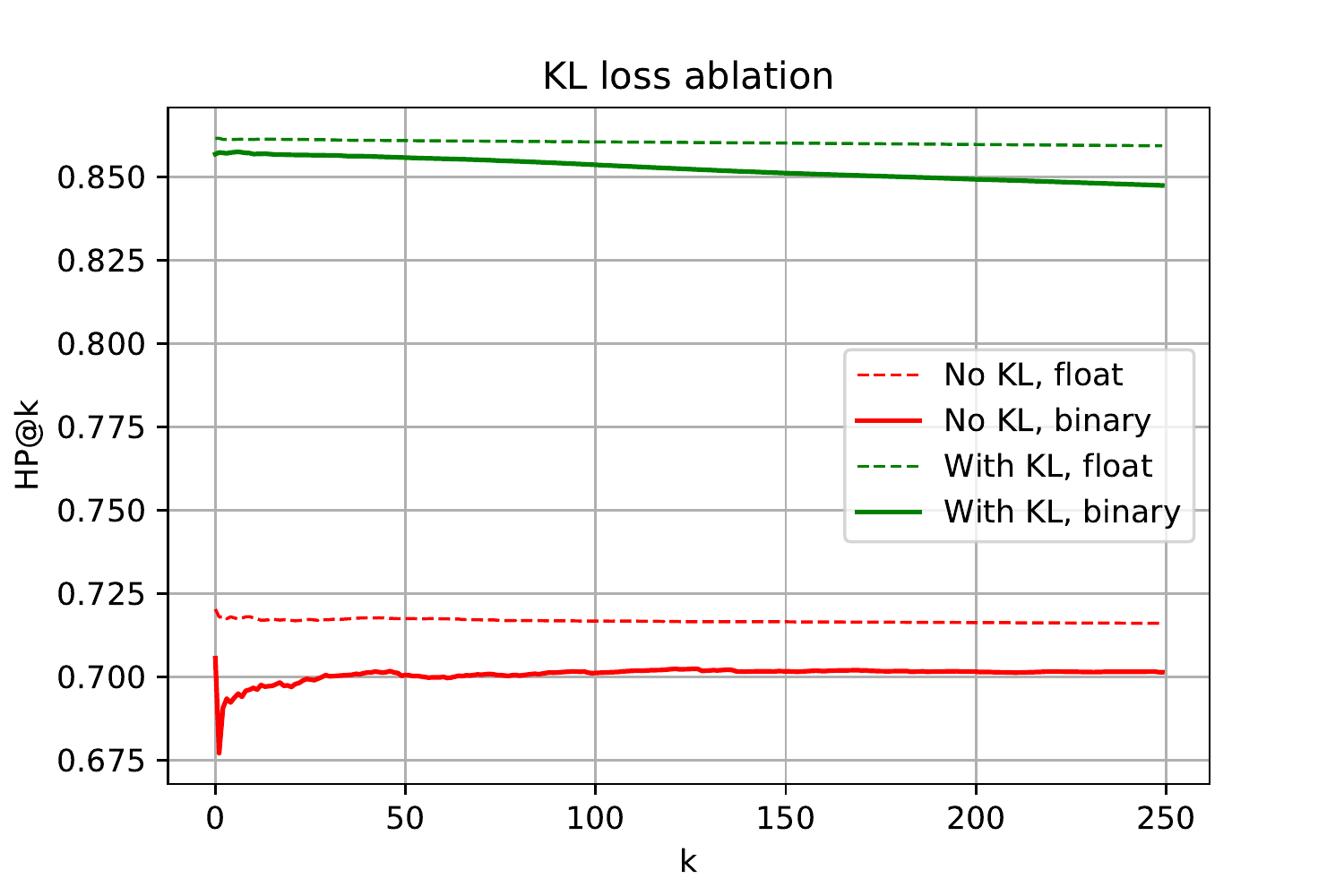}
\includegraphics[keepaspectratio=true,scale=0.35]{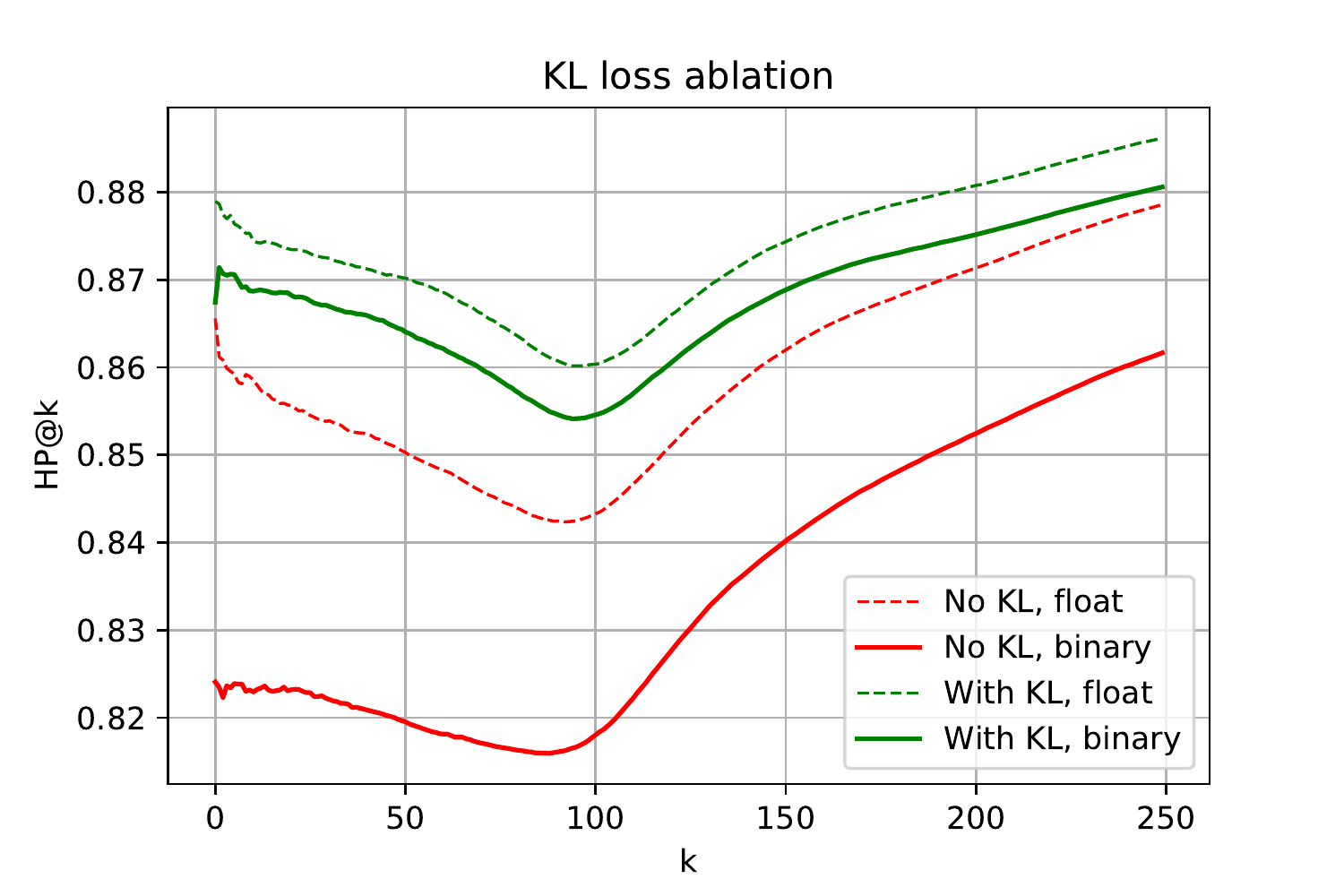}
\end{center}
\caption{Hierarchical precision @k for CIFAR-100 trained model (left) and ILSVRC-2012 WUP trained model (right) for 64-bit codes. We see a substantial drop-in the precision after binarization when not using the KL loss. Binarization does not cause as severe a drop in precision when using the KL loss. A random binary baseline result would yield a curve near the value 0.5.
}
\label{HPgraphImagenet}
\end{figure}

\begin{table*}[t]
\begin{center}
\begin{tabular}{lccccc}
\\ \hline
\textbf{Method} & \textbf{mAHP} & \textbf{mAHP} & \textbf{Kendall} & \textbf{Kendall} & \textbf{Binary}   \\ 
& \textbf{@250}     & \textbf{@250 bin}  & \textbf{Tau float} & \textbf{Tau bin}  & \textbf{Entropy} \\ \hline
SIM-KL               & 0.7566  & 0.7426  & 0.2114 & 0.0910 & 19.63     \\ 
REG                  & 0.7767  & 0.7487  & 0.0842 & 0.0609 & 16.13     \\

\hline
\end{tabular}
\caption{Retrieval results for models trained with Conceptual Captions\label{cc_table} (64 bit codes).}
\end{center}
\end{table*}

\subsection{Google's Conceptual Captions}\label{sec:CC}
We use the same modified ResNet50 architecture as above. We use same hyperparameters as with ImageNet, except we use the Adam optimizer with learning rate 3e-4, and stop training at 120 epochs. We average pool the BERT sentence embeddings and use them to compute the distance matrix for each minibatch. There are unfortunately no standard metrics to measure retrieval performance when no class labels are given. We therefore resort to reporting the mAHP and Kendall Tau distance\cite{kendall1938new} by using Manhattan distances to compare the ranking of one million results retrieved per query with the codes vs. the sentence embeddings. The Kendall Tau distance has the advantage that it doesn't depend on the actual distance but only the ranking (although the ranking is still determined by the sentence embeddings), and that it can be compared to the sentence embedding Kendall Tau distance, which is 0.395 for the BERT embeddings.
We compare a SIM-KL model with a ``REG" model trained by regressing to the embeddings, and again observe the benefits of a KL loss.
We also report the Binary Entropy, and again observe significantly lower values for the REG model. 
%
%
%
\subsection{Zero shot hashing results}\label{sec:ZSH}
%
%
We also test our method on Zero Shot Hashing, following the procedure in \cite{shen2019scalable}; see Table~\ref{zsh_table}. Our models are trained with (i) the ILSVRC2012 1000 class dataset, or (ii) the CC dataset, and tested on completely unseen classes by retrieving from the full 14M ImageNet dataset with the ILSVRC2012 labels removed. We see that our models perform very well in a completely out-of-sample retrieval problem, and specifically perform much better then the classification only baseline.

\begin{table}[]
\begin{center}
\resizebox{300pt}{!}{%
\begin{tabular}{l|c|c|c|c|c}
\hline
\multicolumn{1}{l}{\bf Flat hit @K}  &\multicolumn{1}{c}{K=1} &\multicolumn{1}{c}{K=2} &\multicolumn{1}{c}{K=5} &\multicolumn{1}{c}{K=10} &\multicolumn{1}{c}{K=20}
\\ \hline
ConSE \cite{norouzi2013zero}  &  1.3  & 2.1   & 3.8   & 5.8   & 8.7  \\
SynC \cite{changpinyo2016synthesized}  &  1.5  & 2.4    & 4.5   & 7.1   & 10.9  \\
EXEM (1NNs) \cite{changpinyo2017predicting} & 1.8   & 2.9   & 5.3   & 8.2   & \textbf{12.2}  \\
BZSL \cite{shen2019scalable}  &  1.4  & 2.5    & 4.6   & 7.3   & 11.1  \\ \hline
SIM-KL-CLASS (\textit{Ours, trained on ILSVRC2012})  &  9.9  & 16.0   &  24.6  &  31.7  & 38.9  \\
CLASS (\textit{Ours, trained on ILSVRC2012})   &  15.3  &  20.5  &  28.8  &  34.2  & 41.4  \\
SIM-KL-REG (\textit{Ours, trained on CC dataset})   & 11.1   &  13.2  & 16.8   &  20.3   & 24.5   \\ \hline
\end{tabular}%
}
\caption{The Flat hit@K returning the percentage of test images for which the model returns the one true label in its top K predictions. Four first results are from \cite{shen2019scalable}. SIM-KL-CLASS is our ImageNet trained 64-d model using the full $L_{sim}$, $L_{KL}$, and $L_{cls}$ Losses. CLASS is with only $L_{cls}$. SIM-KL-REG is the model trained with the CC dataset, using also a regression loss towards the BERT sentence embeddings.}
\label{zsh_table}
\end{center}
\end{table}

\subsection{Example retrieval results}
\begin{figure}[h]
\begin{center}
    \includegraphics[keepaspectratio=true,scale=0.23]{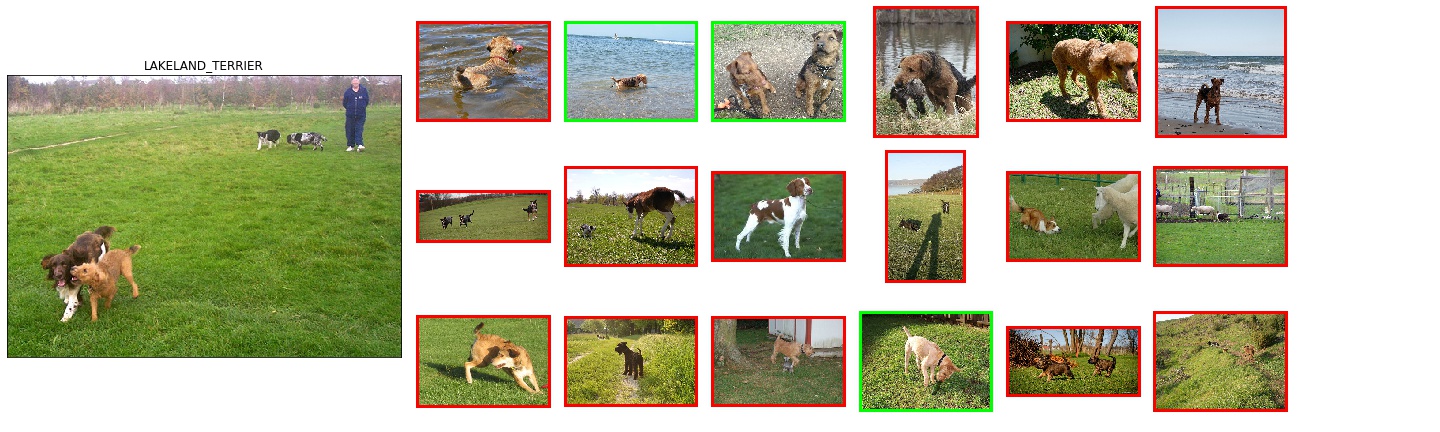}
    \vspace{-10pt}
\end{center}
\caption{Nearest neighbor retrieval results for the large query image on the left, by 3 models on ImageNet dataset. Top row results are retrieved from the 64 bit hash codes of an ImageNet-trained resnet50 model; middle row from a Conceptual Captions (CC)-trained model; bottom row from a concatenation of the two models' codes. We see that although the CC model doesn't retrieve the same class label (probably the captions are not very specific with dog breeds), it does also pay attention the context of the grass in the background and the action of the dogs.\label{ImageNetRetrieval1}}
\end{figure}

\begin{figure}[h]
\begin{center}
    \includegraphics[keepaspectratio=true,scale=0.23]{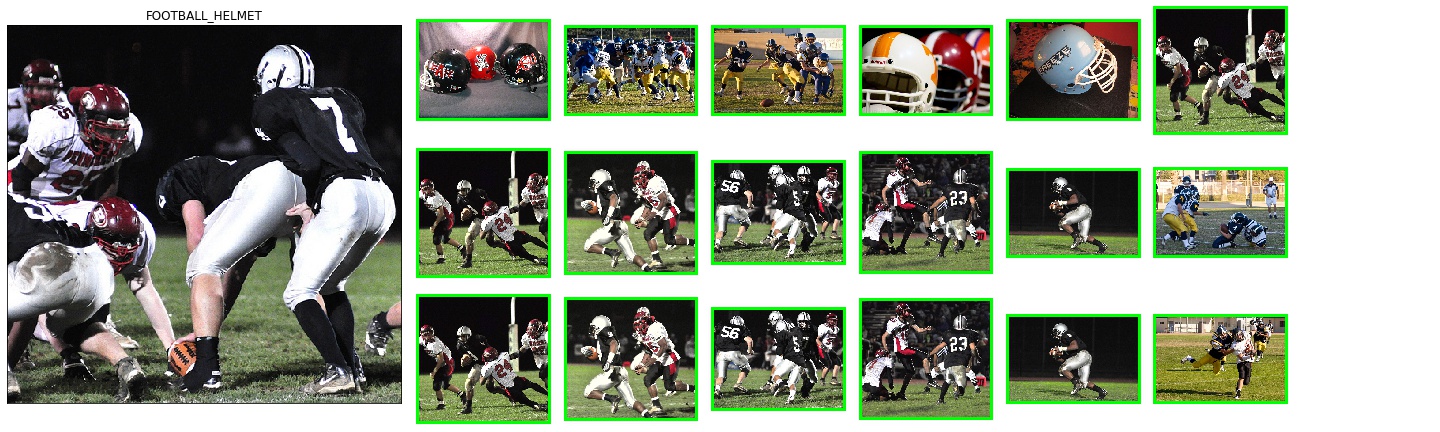}
    \vspace{-10pt}
\end{center}
\caption{The ImageNet-trained model retrieves some football helmets and some footballer players with helmets, but the CC-trained model focuses on the action in a similar way to the query image.\label{ImageNetRetrieval2}}
\end{figure}

See Figs.~\ref{ImageNetRetrieval1} and \ref{ImageNetRetrieval2} for example retrieval results on the ImageNet dataset, and Fig.~\ref{FirstCCfig} for the CC dataset. More examples are shown in the supplementary material.

\begin{figure}[h]
\begin{center}
    \includegraphics[keepaspectratio=true,scale=0.25]{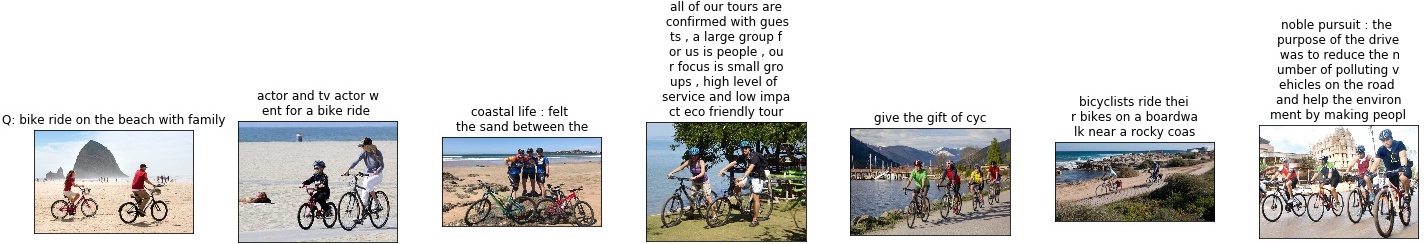}\\
    \vspace{-0pt}
    \vspace{-0pt}
    \includegraphics[keepaspectratio=true,scale=0.25]{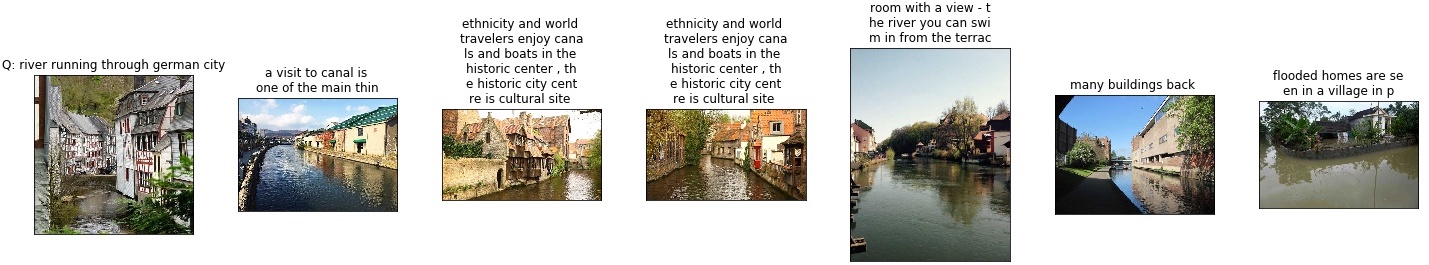}
\end{center}
\vspace{-5pt}
\caption{CC dataset retrieval, using CC-trained model. First image on the left is the query image, others top retrieved results. Note only the image content is used, in retrieval, but original captions are shown for info; the retrieved images' content is more similar to the query than their respective captions. Notice that the CC model has learned to attend to several aspects of similarity including the overall setting/context of the scene and the actions of the people.\label{FirstCCfig}}
\end{figure}

\section{Conclusions}

We presented a novel method to learn binary codes representing semantic image similarity by defining a distance matrix per minibatch of samples, and training the network to learn to match these distances. 
We also showed that by using the empirical KL loss, information loss can be minimized when the continuous valued codes are quantized. This leads to virtually no decrease in retrieval performance when using hashing based retrieval methods, suitable for efficient semantic retrieval on massive databases. Without this loss, performance can be degraded significantly. 
We also showed how modern language models can be used to extract semantically similar caption embeddings, which can be used in the semantic learning to hash scheme. 
Another interesting result is that such language models also yield higher quality learned embeddings on class based data with ImageNet, than using the native WordNet-based WUP measures. We have presented the Kendall-Tau metric as one suggested metric when no classes are available. 
We have also demonstrated real world retrieval performance on unseen classes, and learning detailed notions of semantic similarity beyond class labels.

\appendix
\section{Appendix}

\subsection{Selection criteria for caption embeddings}
We use the STSbenchmark sentence similarity dataset \cite{cer2017semeval} to select embeddings best suited for our purposes, i.e. the embeddings will capture semantic similarity of captions as accurately as possible. Each sentence pair in the dataset is scored from 0 to 5 according to semantic similarity. We extract average pooled embeddings for various Transformer based sentence encoders from the huggingface repository \cite{Wolf2019HuggingFacesTS} for each sentence pair, and compare the Spearman correlation with the ground truth similarity. We also sort the sentence pairs by embedding based predicted similarity, and compare the ranking by Kendall Tau distance. Both scoring methods imply that the BERT model \cite{devlin2018bert} finetuned with the MRPC dataset \cite{dolan2004unsupervised} is the best at capturing sentence semantic similarity, although it is possible that other methods than simple average pooling could yield even better results (some results, such as with the RoBERTa yielded suspiciously poor results). 
Further details of these scores in shown the supplementary material.


\clearpage
%
%
\bibliographystyle{splncs04}
\bibliography{main}
\end{document}